\documentclass[letterpaper]{article} 
\usepackage{aaai23}  
\usepackage{times}  
\usepackage{helvet}  
\usepackage{courier}  
\usepackage[hyphens]{url}  
\usepackage{graphicx} 
\usepackage{natbib}  
\usepackage{caption} 
\usepackage{algorithm}
\usepackage{algorithmic}

\usepackage{newfloat}
\usepackage{listings}
\DeclareCaptionStyle{ruled}{labelfont=normalfont,labelsep=colon,strut=off} 
\floatstyle{ruled}
\newfloat{listing}{tb}{lst}{}
\floatname{listing}{Listing}
\usepackage{multirow}

\usepackage{amsmath,amssymb}
\usepackage{mathtools}
\usepackage{subcaption}

\usepackage{subcaption}

\usepackage{color}
\usepackage[dvipsnames]{xcolor}
\usepackage{xcolor}

\def\etal{\textit{et al}. }
\def\ie{\textit{i.e. }}

\def\vs{\textit{v.s. }}

\colorlet{CLRBlue}{black}

\newcommand{\wdairv}[1]{{\color{CLRBlue}{#1}}}

\title{Semi-Supervised Deep Regression with Uncertainty Consistency and \\ Variational Model Ensembling via Bayesian Neural Networks}
\author{
    Weihang Dai\textsuperscript{\rm 1}, Xiaomeng Li\textsuperscript{\rm 1, \rm 2}\thanks{Corresponding author}, Kwang-Ting Cheng\textsuperscript{\rm 1, \rm 2}
}
\affiliations{

    \textsuperscript{\rm 1} Department of Computer Science and Engineering, The Hong Kong University of Science and Technology \\
    \textsuperscript{\rm 2} Department of Electronic and Computer Engineering, The Hong Kong University of Science and Technology \\
    wdaiaj@connect.ust.hk; eexmli@ust.hk; timcheng@ust.hk;
}

\begin{document}

\maketitle

\begin{abstract}
Deep regression is an important problem with numerous applications. These range from computer vision tasks such as age estimation from photographs, to medical tasks such as ejection fraction estimation from echocardiograms for disease tracking.
Semi-supervised approaches for deep regression are notably under-explored compared to classification and segmentation tasks, however. 
Unlike classification tasks, which rely on thresholding functions for generating class pseudo-labels, regression tasks use real number target predictions directly as pseudo-labels, making them more sensitive to prediction quality.
In this work, we propose a novel approach to semi-supervised regression, namely Uncertainty-Consistent Variational Model Ensembling (UCVME), which
improves training by generating high-quality pseudo-labels and uncertainty estimates for heteroscedastic regression.
Given that aleatoric uncertainty is only dependent on input data by definition and should be equal for the same inputs, we present a novel uncertainty consistency loss for co-trained models.
Our consistency loss significantly improves uncertainty estimates and allows higher quality pseudo-labels to be assigned greater importance under heteroscedastic regression. 
Furthermore, we introduce a novel variational model ensembling approach to reduce prediction noise and generate more robust pseudo-labels. We analytically show our method generates higher quality targets for unlabeled data and further improves training. 
Experiments show that our method outperforms state-of-the-art alternatives on different tasks and can be competitive with supervised methods that use full labels~\footnote{Code is available at https://github.com/xmed-lab/UCVME}. 

\end{abstract}

\section{Introduction}

Deep learning has achieved state-of-the-art results on a variety of tasks such as classification \cite{dosovitskiy2020image}, segmentation \cite{chen2021semi}, image generation \cite{bodla2018semi}, and others. These methods tend to require large amounts of labeled data for training, however, which can be costly to annotate. State-of-the-art image classifiers such as ViT are trained on the JFT-300M dataset, for example, which consists of 300 million images~\cite{dosovitskiy2020image}. 
Labeling can also be prohibitively expensive for medical image analysis, where life-saving tasks such as medical disease diagnoses \cite{li2021rotation} and tumor segmentation \cite{li2018h} require domain expertise. The ability to train neural networks with reduced labels is therefore highly valuable and an active research area.

\begin{figure}%
\centering

\includegraphics[width=0.99 \columnwidth]{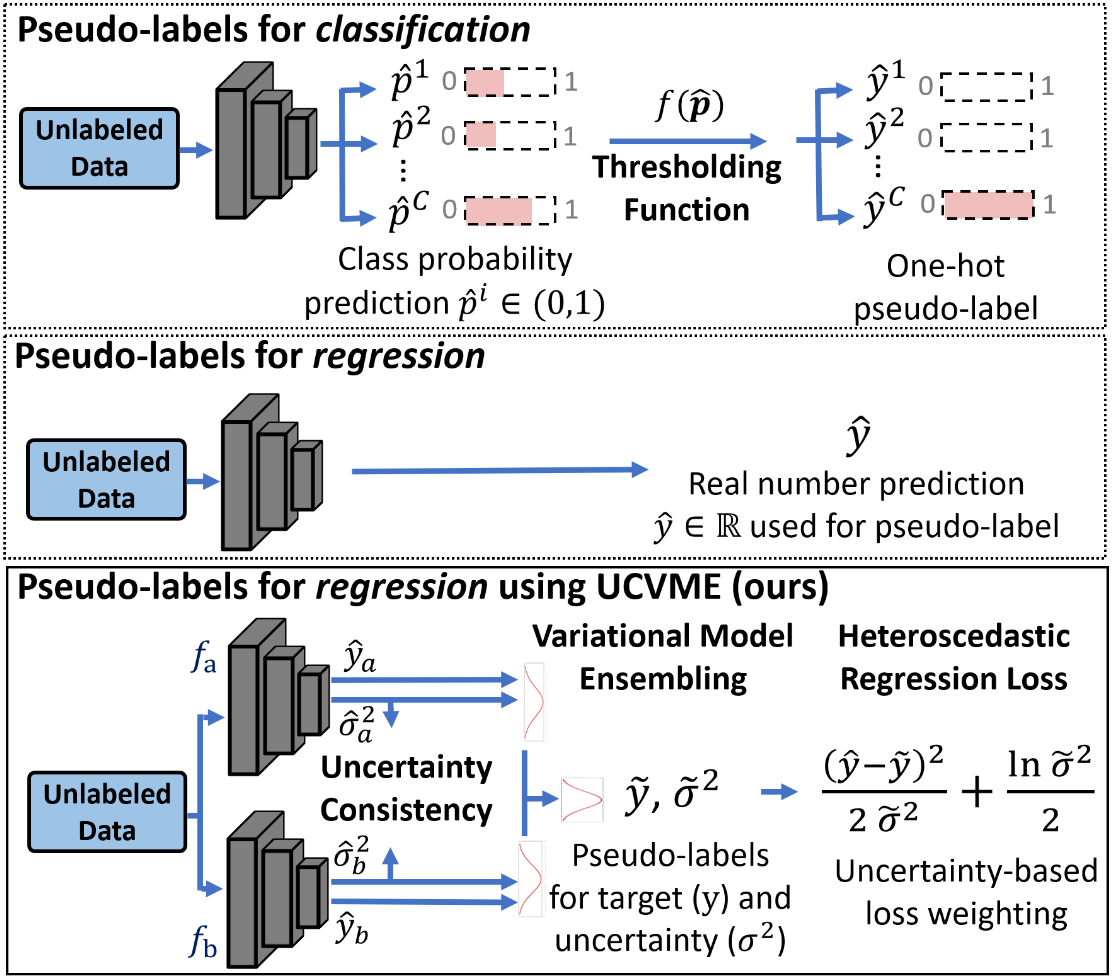}%

\caption{ Differences between standard pseudo-labeling approaches and our method (UCVME). Classification tasks typically apply thresholding functions to probability predictions, \wdairv{$\hat{p}^i$}, to obtain one-hot pseudo-labels, \wdairv{$\hat{y}^i$}. Regression tasks use real number target predictions, \wdairv{$\hat{y}$, directly as pseudo-labels} and are therefore more sensitive to prediction quality. 
Our UCVME improves pseudo-labels for regression by considering pseudo-label uncertainty, \wdairv{$\sigma ^ 2$}, and robustness. We use a novel uncertainty consistency loss to improve uncertainty-based loss weighting and a variational model
ensembling method to improve pseudo-label quality.
}
\label{intro_pl}
\end{figure}




Semi-supervised learning uses unlabeled data together with a smaller labeled dataset for model training. These methods reduce reliance on labeled data and sometimes outperform state-of-the-art techniques on fully labeled datasets. Chen \etal{} \cite{chen2021semi} propose CPS, a semi-supervised algorithm for image segmentation, which is time-consuming to label. Li \etal{} \cite{li2021rotation} enforce consistency between transformed inputs for medical diagnosis, which requires specialist knowledge for annotation. 
However, comparatively less attention has been paid to deep regression problems, which cover practical applications such as age estimation \cite{berg2021deep} and pose estimation  \cite{yang2019fsa} from images. Deep regression is particularly important in the medical field as it is used to obtain measurements for disease diagnosis and progression tracking, such as bone mineral density estimation for osteoporosis~\cite{hsieh2021automated} and ejection fraction estimation for cardiomyopathy~\cite{ouyang2020video}. 

Regression problems are fundamentally different from classification problems because they generate real number predictions instead of class probabilities. 
Existing semi-supervised classification techniques \textit{cannot be applied to semi-supervised regression} because they rely on class probabilities and thresholding functions to generate pseudo-labels~\cite{zhang2021flexmatch, sohn2020fixmatch} (see Fig. \ref{intro_pl}).
Limited efforts have been devoted to exploring semi-supervised approaches for deep regression. Recent works by Jean \etal{}~\cite{jean2018semi} propose deep kernel learning for semi-supervised regression, but their method is designed for tabular data. Pretrained feature extractors are used for image inputs, which prevents task-specific feature learning and limits performance. Wetzel \etal{} \cite{wetzel2021twin} propose TNNR, which estimates the difference between inputs with deep networks and uses loop consistency for unlabeled data. Loop consistency regulates training, but poor-quality predictions can still reduce the effectiveness of the constraints (see Tables \ref{table:sota_age} and \ref{table:sota_ef}).

Unlike classification tasks, which can smooth predictions using thresholding functions for class pseudo-labeling, regression tasks directly use real number target predictions as pseudo-labels.
Therefore, model performance highly depends on the quality of pseudo-labels, \ie predictions. 
In this paper, we propose a novel Uncertainty-Consistent Variational Model Ensembling method, namely UCVME, that adjusts for the uncertainty of pseudo-labels during training and increases pseudo-label robustness. Our method is based on two key ideas: enforcing uncertainty consistency between co-trained models to improve uncertainty-based loss weighting, and using ensembling techniques to reduce prediction variance for obtaining higher quality pseudo-labels.

We make use of Bayesian neural networks (BNNs), which predict aleatoric uncertainty of observations jointly with the target value. The uncertainty estimates are used for heteroscedastic regression, which assigns sample weightings based on uncertainty to reduce the impact of noisier samples~\cite{kendall2017uncertainties}. 
We observe that aleatoric uncertainty, which by definition is dependant only on input data, \textit{should be equal for the same input,} and propose a novel consistency loss for uncertainty predictions of co-trained models. 
Our proposed loss notably improves aleatoric uncertainty estimates on unlabeled data, such that higher quality pseudo-labels are given greater importance through heteroscedastic regression 
(see Fig. \ref{aleac_plot_age}).
\wdairv{This is non-trivial since unreliable uncertainty estimates can lead to adverse loss-weighting and unstable training. Our proposed method is \textit{the first to address uncertainty estimation quality for regression}}.

\begin{figure}%
\centering

\includegraphics[width=0.99 \columnwidth]{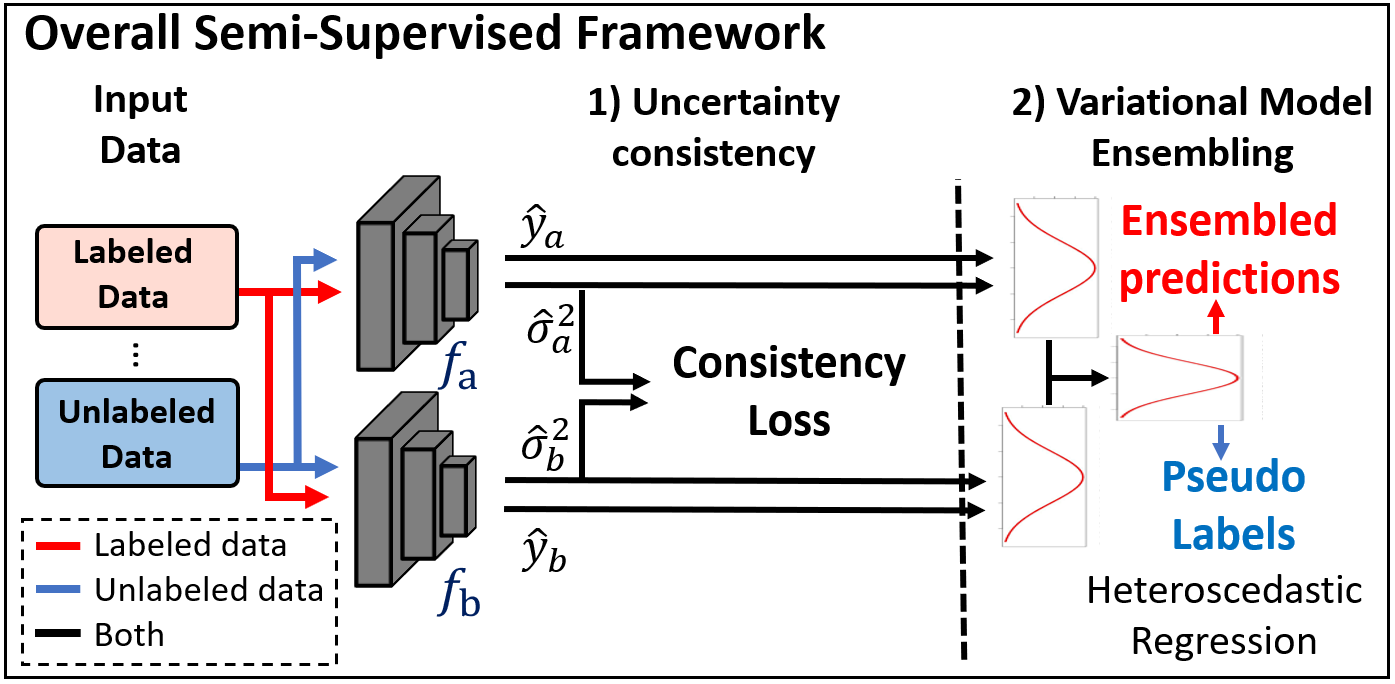}%

\caption{\wdairv{Semi-supervised deep regression framework for our UCVME method. UCVME} improves overall pseudo-label quality and assigns greater sample weights to pseudo-labels with low uncertainty. 
}
\label{intro}
\end{figure}

BNNs also use variational inference during prediction to approximate the underlying distribution of estimates. 
To improve robustness of pseudo-labels, we introduce variational model ensembling, which uses ensembling methods with variational inference to reduce prediction noise. We analytically show our approach generates higher quality targets for unlabeled data and validate results experimentally (see Table \ref{table:abl_age}). 
The combined improvements in uncertainty estimation and pseudo-label quality lead to state-of-the-art performance. 
Fig.~\ref{intro} illustrates the overall framework. 

We demonstrate our method on two regression tasks: age estimation from photographs and ejection fraction estimation from echocardiogram videos.
Results show our method outperforms state-of-the-art alternatives and is competitive with supervised approaches using full labels (see Tables \ref{table:sota_age} and \ref{table:sota_ef}). Ablations demonstrate individual contributions from uncertainty consistency and variational model ensembling (Table \ref{table:abl_age}). We summarize our main contributions as follows:

\begin{itemize}
    \item 
    We propose UCVME, a novel semi-supervised method that improves uncertainty estimates and pseudo-label robustness for deep regression tasks. 
    
    \item 
    We introduce a novel consistency loss for aleatoric uncertainty predictions of co-trained models, based on the insight that estimates should be equal for the same input. 
    
    \item 
    We introduce variational model ensembling 
    for generating pseudo-labels on unlabeled data, which we analytically show is more accurate than deterministic methods. 
    
    \item
    Results show our method outperforms existing state-of-the-art alternatives on two separate regression tasks. 

\end{itemize}

\section{Related Works}
In this section, we review works on learning from unlabeled data, general approaches for semi-supervised learning, state-of-the-art methods for semi-supervised regression, \wdairv{and existing methods for uncertainty estimation}.

\subsection{Unsupervised Representation Learning }
One way to learn from unlabeled data is to learn unsupervised feature representations, which can then be fine-tuned for specific tasks using classifiers.
Techniques such as PCA~\cite{bengio2013representation} and data clustering~\cite{huang2016unsupervised} learn intermediate features by reducing input dimensionality.
With the increasing effectiveness of deep learning, pre-text tasks such as input reconstruction \cite{kingma2013auto}, augmentation prediction \cite{zhang2019aet}, and order prediction \cite{noroozi2016unsupervised} have been explored for unsupervised training of deep feature extractors. 
Current state-of-the-art approaches are based on contrastive learning, which has been shown in some cases to outperform supervised learning \cite{chen2021empirical,chen2020simple}. 

\subsection{Semi-Supervised Learning}
Semi-supervised learning uses both labeled and unlabeled data for training. This reflects realistic settings where raw data is easy to obtain but annotations can be costly. State-of-the-art methods include enforcing consistency on augmented inputs and using pseudo-labels for unlabeled samples. For example, CCT \cite{ouali2020semi} applies prediction consistency after perturbing intermediate features.
CPS \cite{chen2021semi} enforces consistency of segmentation predictions between co-trained models.
\wdairv{Temporal ensembling \cite{laine2016temporal} and mean-teacher \cite{tarvainen2017mean} methods use prediction and model-weight ensembling respectively to generate pseudo-labels.} FixMatch~\cite{sohn2020fixmatch} and FlexMatch~\cite{zhang2021flexmatch} use class probability thresholding for pseudo-labeling to achieve state-of-the-art results on semi-supervised classification. 
Similar techniques have been applied to video action recognition \cite{xu2021cross}, image generation \cite{bodla2018semi}\wdairv{, medical image segmentation \cite{li2020transformation,li2018semi,you2022momentum,lin2022calibrating}}, and other tasks. 


\subsection{Semi-Supervised Regression}



Regression problems are fundamentally different from classification as they involve predicting real numbers in ${\rm I\!R}$ instead of class probabilities. Semi-supervised classification methods, which use thresholding functions to select high-probability class pseudo-labels \cite{sohn2020fixmatch,zhang2021flexmatch}, cannot be adapted to regression tasks \textit{because there is no equivalent to probability thresholding for real number predictions.} Different formulations must be used instead to quantify prediction uncertainty for regression tasks. 

Less attention has been paid to semi-supervised deep regression despite its \wdairv{importance \cite{jean2018semi,wetzel2021twin, yin2022fishermatch}. 
Semi-supervised regression is especially valuable }in medical image analysis, since regression tasks are widely used and annotation costs are high\wdairv{~\cite{ouyang2020video,hsieh2021automated,dai2022cyclical}}.
COREG~\cite{zhou2005semi} is a semi-supervised regression technique originally proposed in 2005 but still commonly used today. Two KNN regressors are co-trained and used to generate pseudo-labels for unlabeled data. Co-training schemes have also been extended to support vector regression by Xu \etal{}~\cite{xu2011semi}. Graph-based methods proposed in \cite{timilsina2021semi} make use of input proximity for pseudo-labeling. 
More recent works by Jean \etal{}~\cite{jean2018semi} and Mallick \etal{}~\cite{mallick2021deep} make use of deep kernel learning for regression. 

One major disadvantage of these methods is that they are primarily designed for structured inputs, where samples consist of one-dimensional tabular data. Feature extractors cannot be trained end-to-end for unstructured inputs such as images and video. Jean \etal{}~\cite{jean2018semi} for example rely on feature extractors pretrained on ImageNet~\cite{deng2009imagenet} to obtain one dimensional embeddings from images. This limits performance as task-specific features cannot be learned (see results in Tables \ref{table:sota_age} and \ref{table:sota_ef}). TNNR \cite{wetzel2021twin} is an alternative method that uses deep networks to predict differences between input pairs. Loop consistency is applied to ensure looped differences sum to zero. Although loop consistency helps regularize training on unlabeled data, inaccurate predictions can limit its effectiveness (see Tables \ref{table:sota_age} and \ref{table:sota_ef}). 


Unlike previous works, we address semi-supervised deep regression by improving uncertainty estimation and pseudo-label quality for real number targets.
Our UCVME method, which proposes a novel uncertainty consistency loss and variational model ensembling, allows training to be focused on high-quality, robust pseudo-label targets and achieves state-of-the-art results on different regression tasks. 

\subsection{\wdairv{Uncertainty Estimation}}

\wdairv{Uncertainty estimation is commonly used in semi-supervised learning to adjust for pseudo-label quality of unlabeled samples.
UA-MT \cite{yu2019uncertainty} and UMCT \cite{xia20203d} both use Monte Carlo dropout to estimate pseudo-label uncertainty, which is then used to filter pseudo-labels or weight unlabeled samples for segmentation tasks.
Yao \etal{} \cite{yao2022enhancing} and Lin \etal{} \cite{lin2022calibrating} estimate uncertainty based on prediction differences between co-trained models for segmentation of medical images.
Semi-supervised classification methods such as FixMatch \cite{sohn2020fixmatch} and FlexMatch \cite{zhang2021flexmatch} implicitly filter out uncertain pseudo-labels by setting confidence thresholds for predictions. 

Uncertainty estimation approaches designed for semi-supervised deep regression have not been explored in existing works however. 
Although methods such as heteroscedastic regression can be used to estimate uncertainty, it can only be done through joint prediction with the target label \cite{kaufman2013heteroskedasticity}. Na\"ive implementation using pseudo-labels give unreliable estimates, which can leads to inaccurate pseudo-labels being assigned larger weights (see Fig.~\ref{aleac_plot_age}). In this work, we propose a novel uncertainty consistency loss that significantly improves the quality of uncertainty estimates on unlabeled data.
This results in more effective uncertainty-based sample weighting and leads to state-of-the-art performance on different semi-supervised deep regression tasks. }

\section{Methodology}





UCVME is based on two novel ideas: enforcing aleatoric uncertainty consistency to improve uncertainty-based loss weighting, and variational model ensembling for generating high-quality pseudo-labels. 
We make use of Bayesian neural networks, which differ from regular neural networks by their usage of aleatoric uncertainty prediction and variational inference \cite{kendall2017uncertainties}. 
We denote $\mathcal{D} \coloneqq \left \{ (x_i, y_i) \right \}_{i=1}^{N}$ as the labeled dataset consisting of $N$ samples, where $x_i$ is the input data and $y_i$ is its corresponding label. 
We denote $\mathcal{D}' \coloneqq \left \{ x'_{i'} \right \}_{i'=1}^{N'}$ as the unlabeled dataset consisting of input data only. 
We train two BNNs, $f_m$ where $m \in \{a,b\}$, in a co-training framework and use Monte Carlo dropout for training and inference.  We denote $\hat{y}_{i,m}$ as model $m$'s prediction for target label $y_i$. 
We denote $\sigma^2_i$ as aleatoric uncertainty but predict log-uncertainty $\ln \sigma^2_i$ in practice, \wdairv{which is always done to avoid obtaining negative predictions for variance. We denote predicted log-uncertainty using $\hat{z}_{i,m}$.}






  \subsection{Aleatoric Uncertainty Consistency Loss for Improved Heteroscedastic Regression}\label{method:ale}
 
 Aleatoric uncertainty, $\sigma^2_i$, refers to uncertainty relating to input data. It is used in BNNs as the variance parameter for heteroscedastic regression loss: 
 
 

\begin{equation} \label{eq:hetloss}
    \mathcal L_{reg} = \frac{1}{N} \sum_{i=0}^N  \frac{(y_i-\hat{y}_{i})^2}{2 \sigma^2_i} + \frac{\ln \sigma^2_i}{2}  \:.
\end{equation}
\wdairv{where $\hat{y}_{i}$ is the prediction for target label $y_{i}$.} Intuitively, the loss function weighs error values dynamically based on aleatoric uncertainty. Samples with high uncertainty are regarded as having lower quality labels with higher noise, and these are given less importance compared to those with greater certainty~\cite{kendall2017uncertainties}. 
Its formal derivation is based on maximum likelihood estimation, assuming observation errors are distributed with different levels of  variance~\cite{kaufman2013heteroskedasticity}. 
In contrast, standard mean squared error (MSE) loss assumes homoscedastic errors, \ie{} uncertainty values $\sigma^2_i$ have equal variance, which is a more restrictive and unrealistic assumption. We refer interested readers to Sup-1 of the supplementary materials for a review of formal derivations and comparisons. 

Heteroscedastic regression can be beneficial for unlabeled data as it allows samples to be weighted based on pseudo-label uncertainty. 
In practice however, uncertainty prediction is difficult because uncertainty has no ground truth label and must be jointly predicted with the target value. 
Unstable predictions that do not reflect label quality can adversely affect training by assigning noisier samples with larger weights.
\textit{Stable training is even more difficult for unlabeled data} because the target ground truth value is also unavailable, which is why heteroscedastic regression has not been successfully used in existing semi-supervised works. \wdairv{We show this effect in Fig. \ref{aleac_plot_age}, where we see uncertainty predictions obtained using heteroscedastic regression only can be unreliable. }

We observe that aleatoric uncertainty for the same input data \textit{should be equal by definition} and introduce a novel consistency loss to enforce consistent uncertainty predictions between co-trained models. 
Prediction consistency is known to be an effective regularizer \cite{chen2021semi} and can be applied to both labeled and unlabeled data to improve estimates. 
\wdairv{By ensuring uncertainty predictions from co-trained models are consistent, we provide an extra training signal in addition to joint estimation with the target label, which helps the model learn more reliable predictions. }
For labeled inputs, we introduce consistency loss, $\mathcal L_{unc}^{lb}$:

\begin{equation} \label{eq:sup_al}
     \mathcal L_{unc}^{lb} =  \frac{1}{N} \sum_{i=1}^N  (\hat{z}_{i,a} - \hat{z}_{i,b})^2 \;,
\end{equation}
which is based on L2 distance. Heteroscedastic regression loss is calculated using the uncertainty predictions:

 \begin{equation} \label{eq:sup_reg}
     \mathcal L_{reg}^{lb} = \frac{1}{N} \sum_{m = a,b} \: \sum_{i=1}^N  \left(  \frac{(\hat{y}_{i,m}-y_i)^2}{2 \exp (\hat{z}_{i,m})} + \frac{\hat{z}_{i,m}}{2}  \right) \; . 
\end{equation}
For unlabeled data, ground truth target labels for $y$ are unavailable, which makes joint uncertainty prediction challenging. We instead make use of variational model ensembling to obtain pseudo-labels for log-uncertainty, $\widetilde{z}_i$, which is used as the training target. We describe variational model ensembling for unlabeled samples in the subsection below.

 \subsection{Variational Model Ensembling for Pseudo-label Generation} \label{method:psuedo}
 
BNNs use Monte Carlo dropout and variational inference to estimate the distribution of the predictor $\hat{y}$. 
To reduce prediction noise, we can use ensembling techniques that reduce predictor variance, which can be demonstrated through bias-variance decomposition. 
The performance of predictor $\hat{y}$ can be evaluated using expected MSE, which we decompose using bias-variance decomposition as follows:

\begin{equation} \label{eq:sngl_bv}
    E[ ( \hat{y}_i - y_i ) ^2] = (E[\hat{y}_i] - y_i)^2 + E[(\hat{y}_i - E[\hat{y}_i])^2] \:,
\end{equation}
where the first right-hand side term is the bias and the second is the variance. If we take individual sample predictions from variational inference, ${\hat{y}_i}^{\:\:t}$, and obtain an ensemble to form a new predictor $\widetilde{y}_i$, we have:

\begin{equation}
    \widetilde{y}_i = \frac{1}{T} \sum_{t=1}^{T} {\hat{y}_i}^{\:\:t} \:,
\end{equation}
where $T$ is the number of samples used. The expected MSE loss of the predictor $\widetilde{y}_i$ is then:

\begin{equation} \label{eq:ensmb_bv}
    E[ ( \widetilde{y}_i - y_i ) ^2] = (E[\widetilde{y}_i] - y_i)^2 + E[(\widetilde{y}_i - E[\widetilde{y}_i])^2] \:.
\end{equation}
The bias terms of the predictors are equal, but the variance term in equation \ref{eq:ensmb_bv} cannot be greater than in equation \ref{eq:sngl_bv} because more samples are observed (see Sup-2 of supplementary materials for more detailed derivations). This means predictor $\widetilde{y}_i$ will have expected MSE lower than or equal to $\hat{y}_i$ and will always have higher quality. 

Based on this effect, we propose variational model ensembling for generating pseudo-labels on both target value $\widetilde{y}_i$ and log aleatoric uncertainty $\widetilde{z}_i$. 
Whereas pseudo-labels for co-trained models typically rely on cross-supervision in state-of-the-art approaches \cite{xu2021cross,chen2021semi}, 
we ensemble the average estimate of the co-trained models and apply variational inference:

\begin{equation} \label{eq:pseudo_target}
     \widetilde{y}_i = \frac{1}{T} \sum_{t=1}^{T} \frac {\hat{y}^{\:\:t}_{i,a} +\hat{y}^{\:\:t}_{i,b} } {2} \:,
\end{equation}

\begin{equation} \label{eq:pseudo_ali}
     \widetilde{z}_i = \frac{1}{T} \sum_{t=1}^{T} \frac { \hat{z}^{\:\:t}_{i,a} + \hat{z}^{\:\:t}_{i,b} } {2} \:.
\end{equation} 
\wdairv{and use this as the pseudo-label for training.}
Compared to cross-supervision, 
pseudo-labels calculated using variational model ensembling \wdairv{are \textit{more accurate because of reduced predictive variance} and} better reflect the true target and uncertainty values. 
 This is especially important for regression targets because pseudo-labels  directly use real number predictions and do not rely on thresholding functions for smoothing. 
 Uncertainty consistency on unlabeled data is then calculated using $\widetilde{z}_i$ as the training target:
 
   \begin{equation} \label{eq:unsup_al}
     \mathcal L_{unc}^{ulb} =  \frac{1}{N'} \sum_{m = a,b} \: \sum_{i=1}^{N'} (\hat{z}_{i,m} - \widetilde{z}_{i})^2 \; .
\end{equation}
Heteroscedastic regression loss for unlabeled data is calculated using $\widetilde{y}_i$ as the target and $\widetilde{z}_i$ as the log-uncertainty:
 
 \begin{equation} \label{eq:usup_reg}
     \mathcal L_{reg}^{ulb} = \frac{1}{N'} \sum_{m = a,b} \: \sum_{i=1}^{N'}  \left(  \frac{(\hat{y}_{i,m}-\widetilde{y}_i)^2}{2 \exp (\widetilde{z}_{i})} + \frac{\widetilde{z}_{i}}{2} \right)\; .
\end{equation}
The improved pseudo-labels lead to more stable heteroscedastic regression, which generates better training signals on unlabeled data. 

\subsection{Overall Semi-Supervised Framework}

 During training, we calculate heteroscedastic regression loss $\mathcal L_{reg}^{lb}$ and aleatoric uncertainty consistency loss $\mathcal L_{unc}^{lb}$ using labeled samples. Pseudo-labels for unlabeled data are generated using Eq. \ref{eq:pseudo_target} and \ref{eq:pseudo_ali} at the start of every training iteration with the most current model weights. The loss values for  $\mathcal L_{reg}^{ulb}$ and $\mathcal L_{unc}^{ulb}$ are calculated for the unlabeled data and jointly optimized with labeled data using the total loss:
  
    \begin{align} \label{eq:loss_all}
     \mathcal L = &  \mathcal L_{reg}^{lb} + \mathcal L_{unc}^{lb} + w_{ulb} \: (\:  \mathcal L_{reg}^{ulb} + \mathcal L_{unc}^{ulb} \: ) \: ,
\end{align}
 where $w_{ulb}$ is the weighting parameter for unlabeled data. Variational model ensembling is also used for test-time inference to obtain $\widetilde{y}_i$ as the final prediction. Pseudo-code is given in S-Algorithm 1 of the supplementary materials.

\section{Experiments}





We demonstrate our method on two semi-supervised deep regression problems: age estimation from photographs and ejection fraction estimation from echocardiogram videos. \footnote{Code is available at https://github.com/xmed-lab/UCVME.}

\subsection{Age Estimation from Photographs}

Age estimation involves predicting a person's age based on their photograph, and is commonly used as a benchmark task for deep regression. 
Facial images can be easily obtained, but accurate age labels may not always be available given concerns over data privacy. Semi-supervised deep regression methods can provide label-efficient approaches for training. 


\begin{figure}%
\centering

\includegraphics[width=0.99\columnwidth]{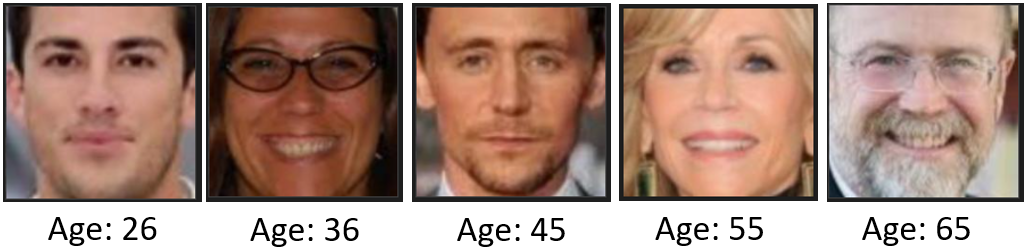}%

\caption{ Sample data from UTKFace dataset \cite{zhang2017age} for age estimation. Pre-cropped images are paired with age labels for training.
}
\label{data_age}
\end{figure}



\setlength{\tabcolsep}{4pt}
\begin{table*}
\fontsize{9pt}{10.8pt}\selectfont
\begin{center}
\caption{Comparison with state-of-the-art methods for age estimation from photographs. We use settings where only 5\%, 10\%, and 20\% of training labels are available. ``Supervised'' methods are only able to use labeled data while ``Semi-supervised'' methods can use labeled and remaining unlabeled data.
``Baseline'' method uses two co-trained BNNs without uncertainty consistency loss and variational model ensembling. 
\textbf{Bold} numbers represent the best result.}
\label{table:sota_age}
\begin{tabular}{l|l|l|ccc|c}

\multicolumn{7}{c}{MAE Values $\downarrow$}\\
\hline
Type & Method & Encoder & 5\% labeled & 10\% labeled & 20\% labeled & All labels \\
\hline
\multirow{1}{*}{Supervised} & RNDB \cite{berg2021deep}    &  ResNet50  & 6.21 $\pm$ 0.12 & 5.69 $\pm$ 0.09 & 5.38 $\pm$ 0.10 & \textbf{4.83 $\pm$ 0.06} \\


\cline{1-7}

\multirow{7}{*}{\begin{tabular}[l]{@{}l@{}}Semi-\\Supervised \end{tabular}} & \wdairv{Mean-teacher \cite{tarvainen2017mean}}  &   ResNet50 & 6.15 $\pm$ 0.08 & 5.54 $\pm$ 0.07 & 5.29 $\pm$ 0.05 & - \\

 & \wdairv{Temporal ensembling \cite{laine2016temporal}}  &   ResNet50 & 6.09 $\pm$ 0.07 & 5.53 $\pm$ 0.05 & 5.25 $\pm$ 0.04 & - \\
 
  & SSDPKL \cite{jean2018semi}  &   ResNet50 & 6.08 $\pm$ 0.06 & 5.50 $\pm$ 0.01 & 5.27 $\pm$ 0.08 & - \\
  
 & TNNR  \cite{wetzel2021twin}  &   ResNet50 & 5.94 $\pm$ 0.04 & 5.41 $\pm$ 0.11 & 5.08 $\pm$ 0.05 & - \\
  & COREG \cite{zhou2005semi}    &   ResNet50 & 5.97 $\pm$ 0.06 & 5.39 $\pm$ 0.04 & 4.97 $\pm$ 0.03 & - \\
  
   & Baseline  &   ResNet50  & 5.92 $\pm$ 0.07 & 5.40 $\pm$ 0.03 & 4.96 $\pm$ 0.03 & - \\
   
& Ours    &   ResNet50 & \textbf{5.84 $\pm$ 0.06} & \textbf{5.26 $\pm$ 0.02} & \textbf{4.85 $\pm$ 0.03} & - \\
\hline
\multicolumn{6}{l}{\textit{ }}\\[-2ex]

\multicolumn{7}{c}{$\mathbf{R}^2$ Values $\uparrow$}\\
\hline

Type & Method & Encoder & 5\% labeled & 10\% labeled & 20\% labeled & All labels \\
\hline
\multirow{1}{*}{Supervised} & RNDB \cite{berg2021deep}     &   ResNet50  & 43.8\% $\pm$ 7.5 & 51.0\% $\pm$ 3.1 & 57.5\% $\pm$ 2.7 & \textbf{65.3\% $\pm$ 0.3} \\
 
\cline{1-7}

\multirow{5}{*}{\begin{tabular}[l]{@{}l@{}}Semi-\\Supervised \end{tabular}}  & \wdairv{Mean-teacher \cite{tarvainen2017mean}}  &   ResNet50 & 45.7\% $\pm$ 1.1 & 54.3\% $\pm$ 0.4 & 58.0\% $\pm$ 0.5 & - \\

 & \wdairv{Temporal ensembling \cite{laine2016temporal}}  &   ResNet50 & 46.1\% $\pm$ 1.0 & 54.2\% $\pm$ 0.4 & 58.9\% $\pm$ 0.3 & - \\
 
  & SSDPKL \cite{jean2018semi}  &   ResNet50 & 46.2\% $\pm$ 1.3 & 54.2\% $\pm$ 0.2 & 58.1\% $\pm$ 0.9 & - \\

 & TNNR  \cite{wetzel2021twin}   &   ResNet50 & 48.6\% $\pm$ 0.3 & 53.1\% $\pm$ 1.5 & 58.6\% $\pm$ 0.5 & - \\
   & COREG \cite{zhou2005semi}    &   ResNet50 & 47.4\% $\pm$ 1.0 & 56.6\% $\pm$ 0.4 & 62.7\% $\pm$ 0.2 & - \\
   
   & Baseline  &   ResNet50   & 47.9\% $\pm$ 1.1 & 56.3\% $\pm$ 0.5 & 62.5\% $\pm$ 0.2 & - \\
   
& Ours    &   ResNet50 & \textbf{49.4\% $\pm$ 0.7} & \textbf{57.9\% $\pm$ 0.3} & \textbf{64.3\% $\pm$ 0.5} & - \\
\hline

\end{tabular}
\end{center}
\end{table*}

\subsubsection{Dataset}


We use the UTKFace dataset~\cite{zhang2017age} and follow the train-test split in previous works~\cite{cao2019rank}. A total of 13,144 images are available for training and 3,287 images for testing. We use a subset of the training dataset for validation. Faces have been pre-cropped and age labels range from 21 to 60 (see Figure \ref{data_age} for examples). 
For our semi-supervised setting, we use subsets of the training data as labeled data and the remaining as unlabeled data. Label distributions are shown in S-Fig. 1 of the supplementary materials.

\subsubsection{Settings} We use ResNet-50~\cite{he2016deep} as our encoder and add additional dropout layers after each of the four main residual blocks. The model is trained for 30 epochs using learning rate $10^{-4}$, weight decay $10^{-3}$, and the Adam optimizer.
We use a batch size of 32 for both labeled and unlabeled data. We set dropout probability as 5\% and use $T=5$ for variational inference. We set $w_{ulb}=10$ which we choose empirically (see S-Table 1 in supplementary materials). 
Mean absolute error (MAE) and $R^2$ are used for evaluation on the test set. Experiments are run six times and mean results are reported with standard deviation.

\subsubsection{Comparison with state-of-the-art}

We compare our method with alternative state-of-the-art approaches for semi-supervised regression, specifically COREG \cite{zhou2005semi},
SSDPKL \cite{mallick2021deep}, and TNNR \cite{wetzel2021twin},
\wdairv{and also adapt mean-teacher \cite{tarvainen2017mean} and temporal ensembling \cite{laine2016temporal} methods for regression.}
\wdairv{To highlight the impact of our proposed components, we introduce} a baseline method \textit{(Baseline)} that uses two co-trained BNNs with heteroscedastic regression loss, but \textit{without} aleatoric uncertainty consistency loss and variational model ensembling. 
We perform training under different semi-supervised settings using only 5\%, 10\%, and 20\% of the available training labels. The remaining samples are treated as unlabeled data. For reference, we \wdairv{also} show results using the supervised state-of-the-art method by Berg \etal{} \cite{berg2021deep} \textit{(RNDB)} for the same settings using reduced labels, as well as on the fully labeled dataset.  

\begin{figure}%
\centering

\includegraphics[width=0.99\columnwidth]{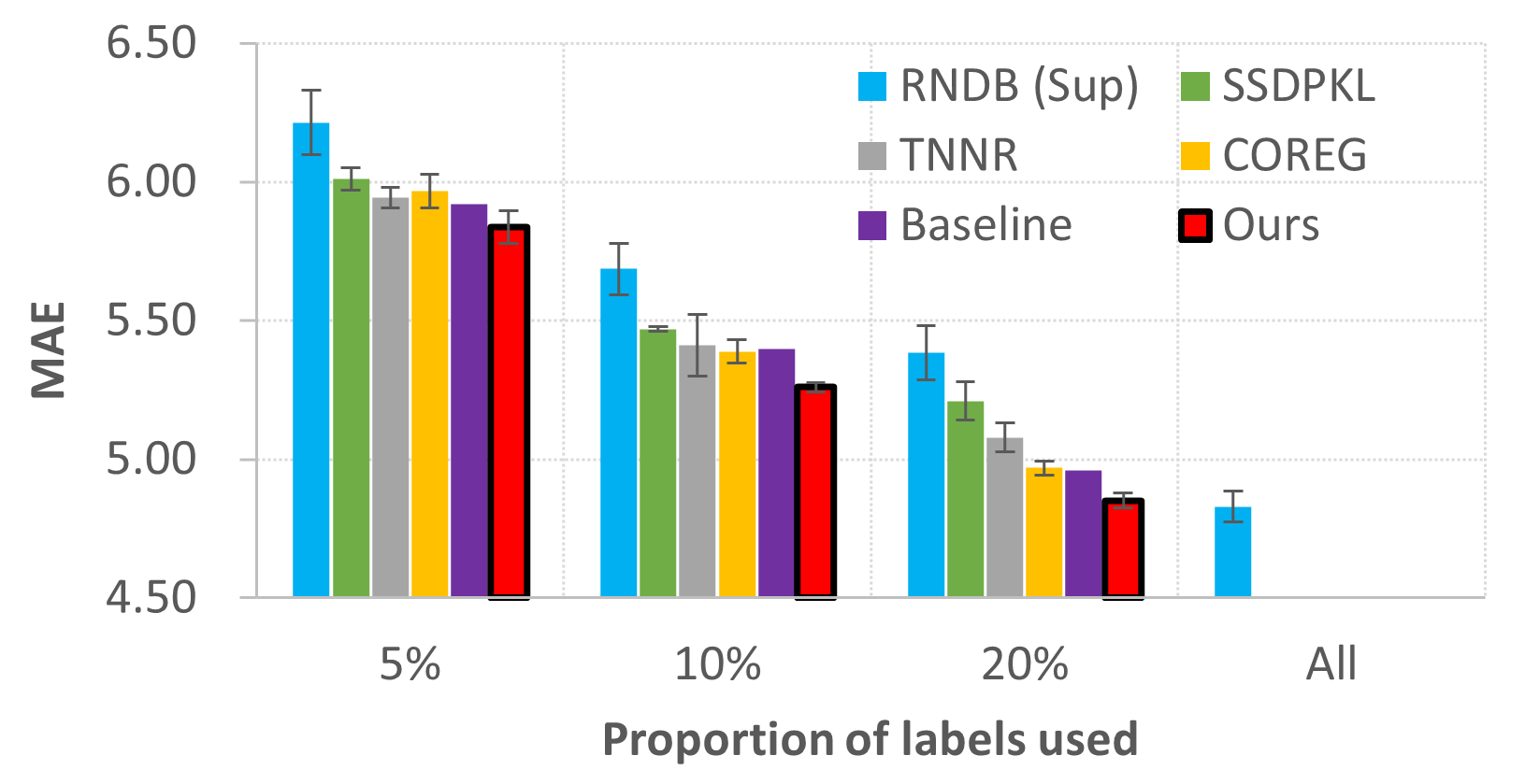}%

\caption{ MAE of different state-of-the-art methods for age estimation. Our proposed method (red) consistently achieves the best results under all settings and is competitive with the supervised approach trained with full labels (blue). 
}
\label{mae_sota}
\end{figure}

The same ResNet-50 encoder~\cite{he2016deep} is used in all methods for fair comparison. We also modify COREG to use co-trained deep regression models instead of KNN regression and use a pretrained feature encoder for 
SSDPKL to obtain image features. Additional implementation details are included in Sup-3 of supplementary materials. We show results in Table \ref{table:sota_age} and visually plot them in Fig. \ref{mae_sota}.

We can see from Fig. \ref{mae_sota} that the supervised approach (blue) under-performs semi-supervised approaches in general. Our method (red) gives the best results and achieves the lowest MAE values for all settings. We also note that our method achieves performance competitive with fully supervised results using only 20\% of available training labels (MAE 4.85 \vs 4.83). UCVME therefore effectively reduces reliance on labeled data for deep regression.





\subsubsection{Ablation Study}

We analyze the performance contribution of different components through ablation. 
We compare results with the baseline model \textit{(Baseline)} after adding uncertainty consistency loss \textit{(Baseline + Con.)}, variational model ensembling \textit{(Baseline + Ens.)}, and both modules
\textit{(Ours)} 
in separate runs to understand the gains from each component. The model is trained with 10\% of the training labels and the rest is used as unlabeled data. Results are shown in Table \ref{table:abl_age}.

 \begin{table}
\fontsize{9pt}{10.8pt}\selectfont
\begin{center}
\caption{Ablation study with 10\% of available labels. Remaining samples are used as unlabeled data. ``Baseline'' method uses two co-trained BNNs without uncertainty consistency loss and variational model ensembling. "Cons." refers to the use of aleatoric uncertainty consistency loss. "Ens." refers to the use of variational model ensembling. 
}
\label{table:abl_age}
\begin{tabular}{l|cc|cc}
\hline
Method & Con. & Ens. &  MAE$\downarrow$ & $\mathbf{R}^2 \uparrow$ \\
\hline

Baseline           & &  & 5.40 $\pm$ 0.03 & 56.3\% $\pm$ 0.5 \\ 
Baseline + Con.            & \checkmark &  & 5.30 $\pm$ 0.01 & 57.6\% $\pm$ 0.2 \\ 
Baseline + Ens.           & & \checkmark  & 5.31 $\pm$ 0.02 & 57.4\% $\pm$ 0.2 \\
Ours    & \checkmark & \checkmark & \textbf{5.26 $\pm$ 0.02}  & \textbf{57.9\% $\pm$ 0.3} \\

\hline
\end{tabular}
\end{center}
\end{table}

We can see consistency loss and variational model ensembling have individual contributions and separately reduce MAE by roughly 0.10. Best results are achieved using both. 

\subsubsection{Impact of consistency loss on uncertainty estimates}

\begin{figure}%
\centering

\includegraphics[width=1\columnwidth]{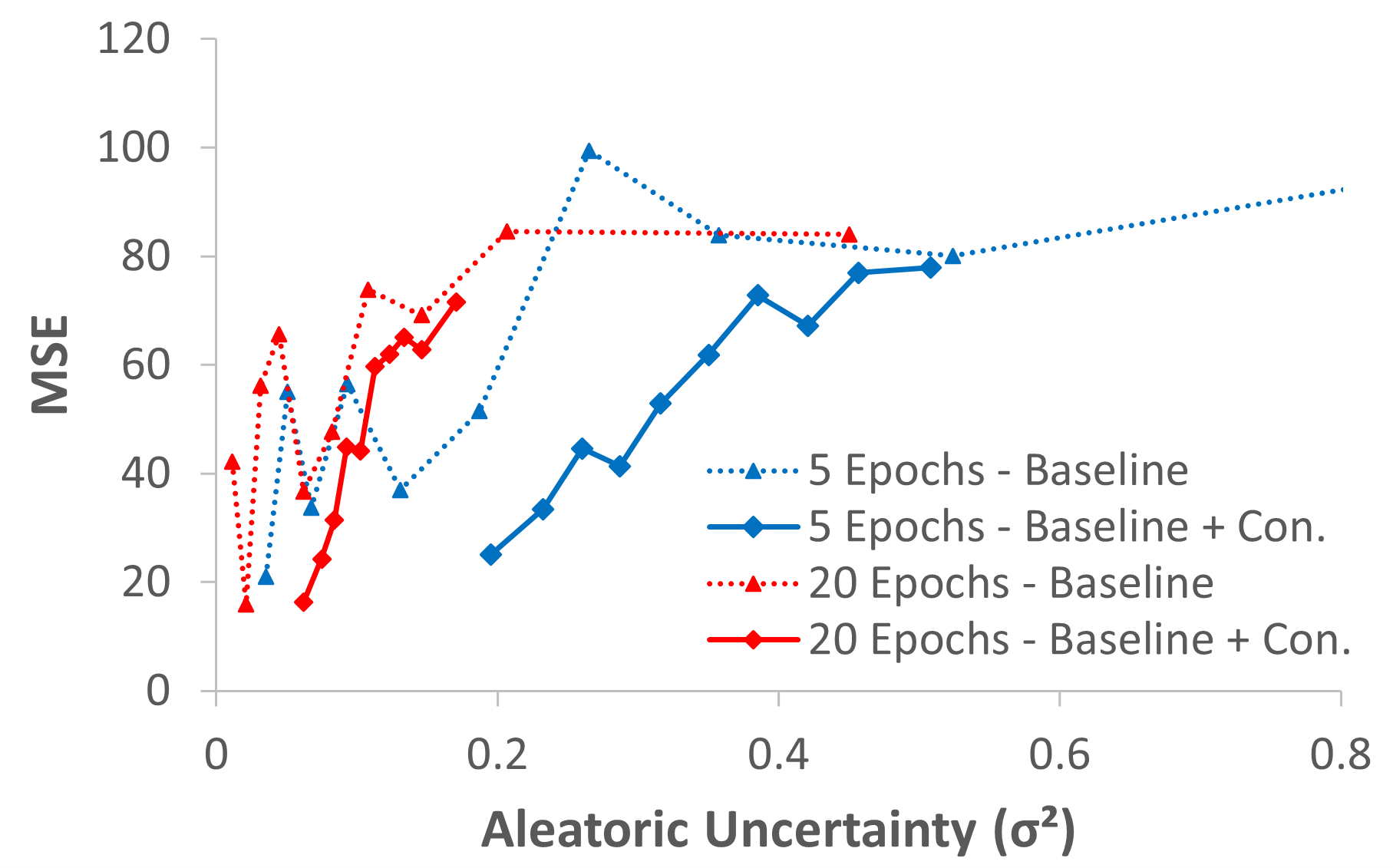}%

\caption{ MSE of pseudo-labels plotted against predicted aleatoric uncertainty. \wdairv{Improved uncertainty estimates should display a stronger negative relationship with quality, since higher uncertainty indicates lower quality pseudo-labels with higher MSE. We can see that uncertainty estimates become more reliable after applying uncertainty consistency loss as the relationship between predicted uncertainty and pseudo-label quality is much clearer (solid \vs dashed lines).} 
}
\label{aleac_plot_age}
\end{figure}

We analyze the impact of consistency loss on uncertainty estimates by visualizing its relationship with pseudo-label quality. Intuitively, improved uncertainty estimates will show a stronger negative relationship with pseudo-label quality, since higher uncertainty means more prediction noise \wdairv{and lower quality labels}. 
We obtain pseudo-labels and uncertainty predictions for unlabeled samples using the \textit{Baseline} and \textit{Baseline + Con.} models. Samples are grouped into ten equal bins based on sorted uncertainty predictions. Pseudo-label quality is measured using MSE against the ground truth target value. Average aleatoric uncertainty is calculated for each group. The two values are plotted against each other in Fig.~\ref{aleac_plot_age} using models trained after five and twenty epochs. 


For the \textit{Baseline + Con.} model, we see overall MSE and uncertainty both decrease after training for more epochs (solid red line \vs solid blue line). Pseudo-labels with lower uncertainty have lower MSE and are of higher quality.
In contrast, uncertainty predictions from the \textit{Baseline} model are more extreme (dashed lines) and are not significantly reduced with training (dashed red line \vs dashed blue line). The relationship between uncertainty and pseudo-label quality is not strong, which can lead to noisier samples being assigned higher loss weightings. Aleatoric uncertainty consistency loss therefore \textit{improves uncertainty estimates significantly and helps prevent adverse sample weighting}.

\subsubsection{\wdairv{Computational Cost}}

\wdairv{
We show in Table \ref{table:compcost} the computational cost of different semi-supervised deep regression approaches in gigaFLOPS per image (G). We note that the cost of UCVME is dependant on $T$, the number of iterations used for variational model ensembling, which is set to 5. For reference, we also show the cost from using a single iteration, $T=1$, which is equivalent to enforcing uncertainty consistency only without using variational model ensembling.

 \begin{table}
\fontsize{9pt}{10.8pt}\selectfont
\begin{center}
\captionsetup{labelfont={color=CLRBlue}}
\caption{ \wdairv{Computation cost of state-of-the-art semi-supervised regression methods in gigaFLOPS per image (G). We also show the cost for our method UCVME without variational model ensembling by setting $T=1$.}
}
\label{table:compcost}

\color{CLRBlue}
\addtolength{\tabcolsep}{-2pt}
\begin{tabular}{l|c}
\hline
\multirow{2}{*}{Method} & Computation \\
& Cost (G)\\
\hline

Mean-teacher \cite{tarvainen2017mean} &  4 \\ 
Temporal ensembling \cite{laine2016temporal} &  4 \\ 
SSDPKL \cite{jean2018semi} &  4 \\
COREG  \cite{wetzel2021twin}    & 21 \\
TNNR  \cite{zhou2005semi}    & 77 \\
\hline
Ours w/o variational model ensembling ($T=1$)      & 21 \\
Ours ($T=5$)     & 49 \\

\hline
\end{tabular}
\addtolength{\tabcolsep}{2pt}
\end{center}
\end{table}

We can see our UCVME method with $T=1$ incurs the same cost as COREG. Regression performance outperforms COREG however due to the use of uncertainty consistency, which can be seen from results in Tables \ref{table:sota_age} and \ref{table:abl_age} (MAE 5.30 \vs 5.39). Using variational model ensembling by setting $T=5$ leads to more computation but better predictions.
Although mean-teacher, temporal ensembling, and SSDPKL require less computation, they do not perform as well.
}

\subsection{Ejection Fraction Estimation from Echocardiogram Videos}

Left ventricular ejection fraction (LVEF) is the most commonly used medical indicator for diagnosing cardiac disease \cite{hughes2021deep}. It is the percentage difference between the maximum and minimum volume of a heart's left ventricle (LV) and measures blood pumping capability. LVEF is manually labeled from echocardiogram video by estimating maximum and minimum volume based on LV segmentations using method of disks \cite{foley2012measuring} and finding their percentage difference. 
An illustration of this process is given in S-Fig. 3 of the supplementary materials for interested readers.  State-of-the-art methods for automating LVEF estimation use spatial-temporal models to perform end-to-end regression on raw video. Video regression requires large amounts of labeled data however, and existing methods by Ouyang \etal \cite{ouyang2020video} and Dai \etal \cite{dai2021adaptive} use up to 10,030 samples for training. Like most medical tasks, annotation requires domain expertise and can be costly, motivating the need for label efficient methods. 

\begin{figure}%
\centering

\includegraphics[width=0.99\columnwidth]{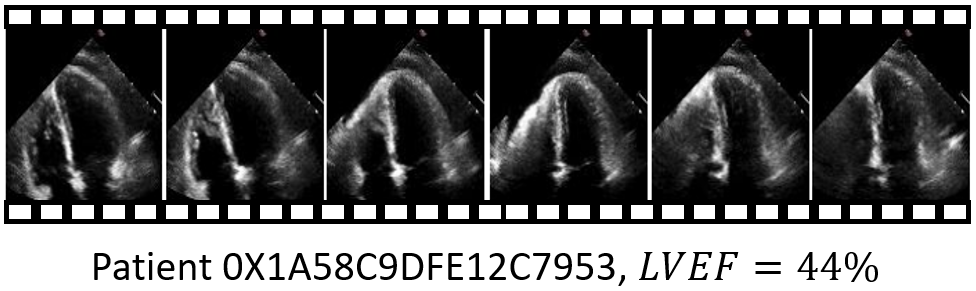}%

\caption{ Sample data from EchoNet-Dynamic \cite{ouyang2019echonet}. Echocardiogram video sequences are paired with LVEF values. 
}
\label{data_ef}
\end{figure}

\begin{table*}
\fontsize{9pt}{10.8pt}\selectfont
\begin{center}
\caption{Comparison with state-of-the-art methods for ejection fraction estimation from echocardiogram video. We use settings where only 1/16, 1/8, and 1/4 of training labels are available. ``Supervised'' methods are only able to use labeled data while ``Semi-supervised'' methods can use labeled and remaining unlabeled data. ``Baseline'' method uses two co-trained BNNs without uncertainty consistency loss and variational model ensembling. \textbf{Bold} numbers represent the best result.}
\label{table:sota_ef}
\begin{tabular}{l|l|l|ccc|c}
\multicolumn{7}{c}{MAE Values $\downarrow$}\\
\hline
Type & Method & Encoder &  1/16 labeled & 1/8 labeled & 1/4 labeled & All labels \\
\hline
\multirow{1}{*}{Supervised} & Ouyang \etal \cite{ouyang2020video}   &   R2+1D  & 6.04 $\pm$ 0.20 & 5.57 $\pm$ 0.21 & 4.78 $\pm$ 0.11 & \textbf{4.13} $\pm$ \textbf{3.85} \\

\cline{1-7}

\multirow{7}{*}{\begin{tabular}[l]{@{}l@{}}Semi-\\Supervised\end{tabular}} & \wdairv{Mean-teacher \cite{tarvainen2017mean}}   &   R2+1D   & 6.01 $\pm$ 0.09 & 5.51 $\pm$ 0.06 & 4.71 $\pm$ 0.07 & - \\

 & \wdairv{Temporal ensembling \cite{laine2016temporal}}   &   R2+1D   & 5.97 $\pm$ 0.08 & 5.52 $\pm$ 0.06 & 4.67 $\pm$ 0.06  & - \\

 & SSDPKL \cite{jean2018semi}   &   R2+1D   & 6.01 $\pm$ 0.04 & 5.47 $\pm$ 0.01 & 4.68 $\pm$ 0.07 & - \\

 & TNNR  \cite{wetzel2021twin}   &   R2+1D  & 5.90 $\pm$ 0.11 & 5.46 $\pm$ 0.08 & 4.79 $\pm$ 0.08 & - \\
  & COREG \cite{zhou2005semi}     &   R2+1D   & 5.94 $\pm$ 0.07 & 5.31 $\pm$ 0.02 & 4.57 $\pm$ 0.02 & - \\
  
 & Baseline  &   R2+1D     & 5.93 $\pm$ 0.10 & 5.36 $\pm$ 0.05 & 4.58 $\pm$ 0.03 & - \\
 
& Ours     &   R2+1D  & \textbf{5.77 $\pm$ 0.04}  & \textbf{5.10 $\pm$ 0.05} & \textbf{4.37 $\pm$ 0.05} & - \\
\hline
\multicolumn{6}{l}{\textit{ }}\\[-2ex]
\multicolumn{7}{c}{$\mathbf{R}^2$ Values $\uparrow$}\\
\hline

Type & Method & Encoder &  1/16 labeled & 1/8 labeled & 1/4 labeled & All labels \\
\hline
\multirow{1}{*}{Supervised} & Ouyang \etal  \cite{ouyang2020video}    &   R2+1D  & 55.3\% $\pm$ 2.6 & 62.5\% $\pm$ 2.2 & 71.6\% $\pm$ 1.4 & \textbf{80.4\%} $\pm$ \textbf{1.2} \\

\cline{1-7}

\multirow{7}{*}{\begin{tabular}[l]{@{}l@{}}Semi-\\Supervised\end{tabular}} & \wdairv{Mean-teacher \cite{tarvainen2017mean}}    &   R2+1D  & 55.1\% $\pm$ 1.4 & 62.9\% $\pm$ 0.7 & 72.5\% $\pm$ 0.4 & - \\

 & \wdairv{Temporal ensembling \cite{laine2016temporal}}    &   R2+1D  & 55.2\% $\pm$ 1.3 & 62.9\% $\pm$ 0.7 & 73.2\% $\pm$ 0.3 & - \\

 & SSDPKL \cite{jean2018semi}    &   R2+1D  & 56.3\% $\pm$ 1.0 & 61.2\% $\pm$ 0.3 & 74.1\% $\pm$ 1.0 & - \\

 & TNNR  \cite{wetzel2021twin}     &   R2+1D  & 55.9\% $\pm$ 1.2 & 63.4\% $\pm$ 0.8 & 73.7\% $\pm$ 0.6 & - \\
 
   & COREG \cite{zhou2005semi}      &   R2+1D  & 55.1\% $\pm$ 0.7 & 64.5\% $\pm$ 0.4 & 74.1\% $\pm$ 0.1 & - \\
   
 & Baseline    &   R2+1D    & 55.2\% $\pm$ 1.4 & 64.9\% $\pm$ 0.3 & 74.5\% $\pm$ 0.1 & - \\

& Ours      &   R2+1D  & \textbf{57.8\% $\pm$ 0.6} & \textbf{66.6\% $\pm$ 0.5} & \textbf{76.3\% $\pm$ 0.6} & - \\
\hline

\end{tabular}
\end{center}
\end{table*}




\subsubsection{Dataset}


We use the EchoNet-Dynamic dataset \cite{ouyang2019echonet}, which consists of 10,030 echocardiogram videos with LVEF labels (see Fig. \ref{data_ef} for examples). The videos have been rescaled to 112 $\times$ 112 pixels and pre-divided into 7,465 videos for training, 1,288 videos for validation, and 1,277 videos for testing. For our semi-supervised setting, we use subsets of the training labels as our labeled data and remaining samples as unlabeled data. Label distributions are given in S-Fig. 2 of the supplementary materials.

\subsubsection{Settings} We use the R2+1D ResNet encoder~\cite{tran2018closer} pretrained on Kinetics 400~\cite{kay2017kinetics} and add additional dropout layers between the four main residual blocks. We set dropout probability as 5\% and use $T=5$ for variational inference. The model is trained using SGD with $10^{-4}$ learning rate and 0.9 momentum for 25 epochs. Learning rate is decayed by 0.1 at epoch 15. Clips of 32 frames are sampled from videos at a rate of 1 in every 2 frames for input. Batches of 10 clips are used for labeled and unlabeled videos. We set $w_{ulb}=10$ which we choose empirically (see S-Table 2 of supplementary materials). We evaluate performance using MAE and $R^2$. Experiments are run five times and mean results with standard deviation are reported.

\subsubsection{Comparison with state-of-the-arts}

\begin{figure}%
\centering

\includegraphics[width=0.99\columnwidth]{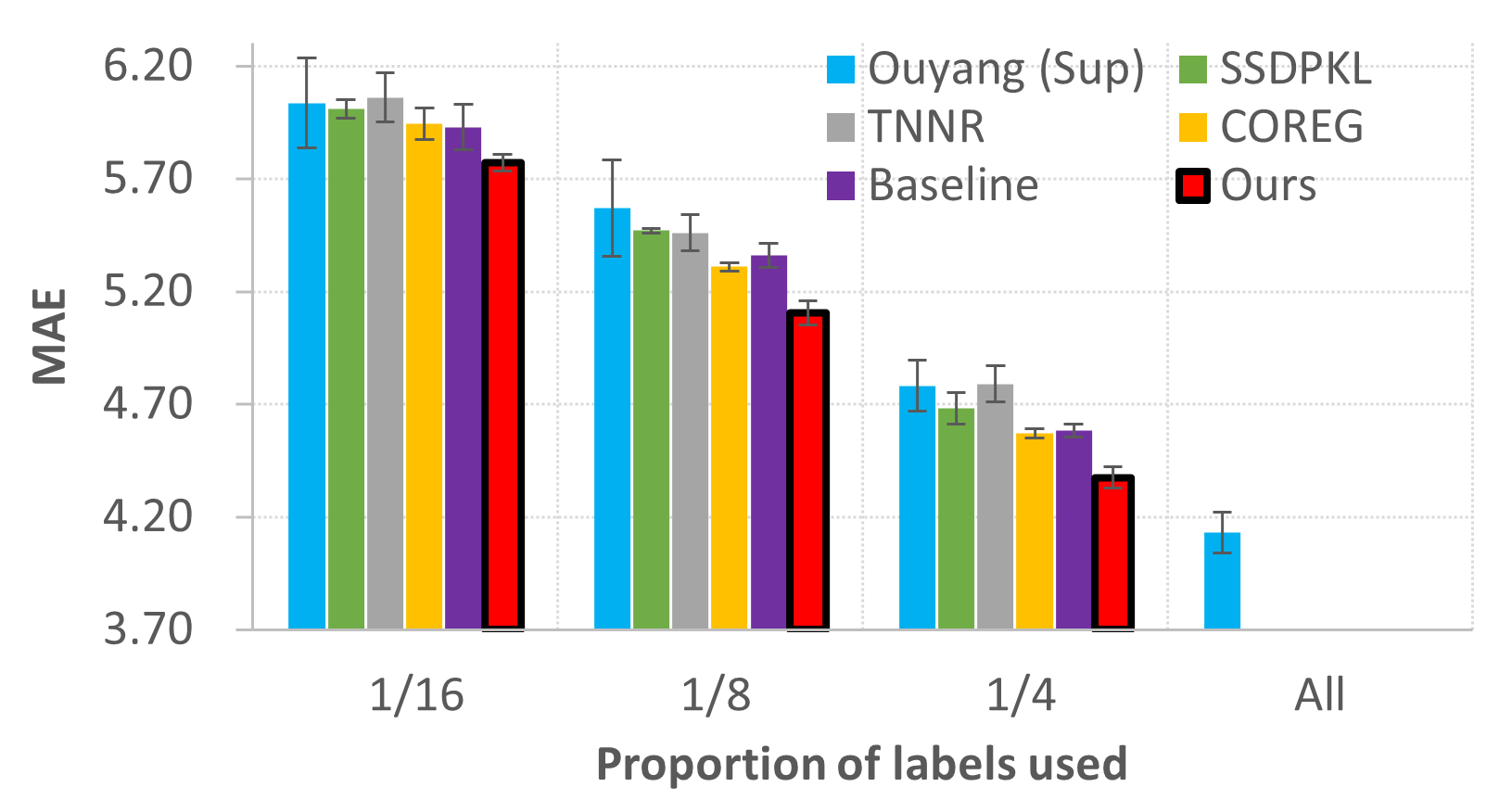}%

\caption{ MAE of different state-of-the-art methods for ejection fraction estimation. Our proposed method (red) consistently outperforms alternatives by significant margins.
}
\label{sota_ef_plot}
\end{figure}

We compare our method with \wdairv{mean-teacher \cite{tarvainen2017mean}, temporal ensembling \cite{laine2016temporal}}, COREG \cite{zhou2005semi},
SSDPKL \cite{mallick2021deep}, TNNR \cite{wetzel2021twin}, and our baseline model \textit{(Baseline)}. 
We perform training under settings where one-sixteenth, one-eighths, and one-quarter of the training labels are used, with the remainder treated as unlabeled data. For reference, we also show results using the supervised method by Ouyang \etal{} \cite{ouyang2020video} on the reduced labels as well as on the fully labeled dataset. 
The Kinetics pretrained R2+1D ResNet encoder \cite{tran2018closer} is used in all methods for fair comparison. Additional implementation details are given in Sup-4 of the supplementary materials. 
Results are shown in Table \ref{table:sota_ef} and plotted in Fig. \ref{sota_ef_plot}. 

Our method consistently achieves the best results for all settings by significant margins. We are also able to achieve an MAE of 4.37 using a quarter of the labels, which is only a relative 5.8\% higher than the 4.13 MAE achieved by Ouyang \etal on fully labeled data. Our proposed method therefore reduces the number of labels required for training which is highly valuable for medical regression tasks.

\section{Conclusion}

In this work, we introduce a novel Uncertainty-Consistent Variational Model Ensembling (UCVME) method for semi-supervised deep regression. Our method improves training on unlabeled data by adjusting for pseudo-label quality and improving pseudo-label robustness. We introduce a novel consistency loss on uncertainty estimates, which we demonstrate significantly improves heteroscedastic loss weighting, especially for unlabeled samples. We also use variational model ensembling to reduce prediction noise and generate better training targets for unlabeled data. 
\wdairv{Our method has strong theoretical support and can be applied to different tasks and datasets. We demonstrate this using two deep regression tasks based on image and video data and achieve state-of-the-art performance for both. Results are also competitive with supervised methods using full labels.} 
UCVME is therefore a valuable method for reducing the amount of labels required for deep regression tasks. 

\section{Acknowledgement}
The work described in this paper was supported by a grant from Hong Kong Research Grants Council General Research Fund (GRF) (16203319), a grant from HKUST-BICI Exploratory Fund under HCIC-004, and a grant from Hong Kong Innovation and Technology
Commission (Project no. ITS/030/21).


\bibliography{aaai23}

\end{document}